\documentclass[10pt,twocolumn,letterpaper]{article}

\usepackage{iccv}
\usepackage{times}
\usepackage{epsfig}
\usepackage{graphicx}
\usepackage{amsmath}
\usepackage{amssymb}

\usepackage{microtype}
\usepackage{subfigure}
\usepackage{booktabs} 
\usepackage{amsthm}
\usepackage{algorithm}
\usepackage{algorithmic}
\usepackage{mathtools}
\usepackage{stfloats}
\usepackage{multirow}
\usepackage{threeparttable}

\usepackage[pagebackref=true,breaklinks=true,letterpaper=true,colorlinks,bookmarks=false]{hyperref}

 \iccvfinalcopy 

\def\iccvPaperID{4342} 

\ificcvfinal\fi
\begin{document}

\title{Co-Evolutionary Compression for Unpaired Image Translation}

\author{Han Shu$^1$, Yunhe Wang$^1$, Xu Jia$^1$, Kai Han$^1$, Hanting Chen$^2$, Chunjing Xu$^1$, Qi Tian$^{1}$\thanks{corresponding author}, Chang Xu$^3$\\
	$^1$Huawei Noah's Ark Lab\\
	$^2$Key Lab of Machine Perception (MOE), CMIC, School of EECS, Peking University, China\\
	$^3$School of Computer Science, Faculty of Engineering, The University of Sydney, Australia\\
	{\tt\footnotesize \{han.shu, yunhe.wang, jiaxu1, kai.han, xuchunjing, tian.qi1\}@huawei.com,} \\
	{\tt\footnotesize htchen@pku.edu.cn, c.xu@sydney.edu.au}
}

\maketitle

	\begin{abstract}
	Generative adversarial networks (GANs) have been successfully used for considerable computer vision tasks, especially the image-to-image translation. However, generators in these networks are of complicated architectures with large number of parameters and huge computational complexities. Existing methods are mainly designed for compressing and speeding-up deep neural networks in the classification task, and cannot be directly applied on GANs for image translation, due to their different objectives and training procedures. To this end, we develop a novel co-evolutionary approach for reducing their memory usage and FLOPs simultaneously. In practice, generators for two image domains are encoded as two populations and synergistically optimized for investigating the most important convolution filters iteratively. Fitness of each individual is calculated using the number of parameters, a discriminator-aware regularization, and the cycle consistency. Extensive experiments conducted on benchmark datasets demonstrate the effectiveness of the proposed method for obtaining compact and effective generators.
\end{abstract}

\section{Introduction}
Generative adversarial networks (GANs~\cite{GAN}) have achieved impressive results in a wide variety of computer vision tasks, such as super-resolution~\cite{GANsr} and image editing~\cite{GANedit}, which are popular applications on mobile devices. Many of these tasks can be considered as an image-to-image translation problem~\cite{pix2pix, pix2pixHD}, where an image from one domain is mapped to a corresponding paired image in the other domain. This task is further extended to the unsupervised learning setting by~\cite{cycleGAN, discoGAN, dualGAN}, where no paired data are required during training. However, launching such image-to-image translation models on mobile devices requires considerable memory and computation cost, which could challenge the hardware performance and would influence users' experience. For instance, it takes about 43\emph{MB} and $1\times10^{10}$ FLOPs (floating-number operations) to process one image of size $224\times224$ using a generator network in the CycleGAN~\cite{cycleGAN}, which requires more resources than some modern CNNs for large-scale  image classification (\eg ResNet~\cite{ResNet} and MobileNet~\cite{Mobilenet}).

Recently, a number of algorithms have been proposed for compressing and speeding-up deep neural networks. For instance, Han~\etal~\cite{pruning} proposed to remove subtle weighs in pre-trained neural networks and rely on some encoding techniques to obtain the compressed models. Wang~\etal~\cite{CNNpackNIPS} further tackled this problem from the perspective of DCT frequency domain to achieve higher compression ratios. 
Luo~\etal~\cite{ThiNet} pruned filters based on statistics information from the next layer. Hu~\etal~\cite{trimming} iteratively pruned the network by removing less important neurons according to the magnitude of feature maps. In addition, there are also several methods proposed for learning portable deep neural networks with different techniques, \eg matrix/tensor decomposition~\cite{SVD}, quantization and binarization~\cite{binary,XNORNet,quan1}, knowledge distillation~\cite{Distill,FitNet,shen2018amal} and efficient convolution blocks design~\cite{Mobilenet,Shufflenet}.

Although the aforementioned methods have made tremendous progress in reducing redundancy in deep neural networks, most of them are designed for recognition tasks such as image classification or object detection. For recognition tasks, neurons with large activation contribute more to the final classification accuracy. Therefore, neurons with weak activation are often eliminated or approximately represented using low-bit data through low-rank decomposition or clustering without obviously degrading the original performance. In contrast, the generative adversarial networks for image translation tasks are usually composed of a generator and a discriminator, which are updated alternatively with a two-player competition strategy. The training of GANs is thus, more difficult than those of conventional neural networks. It is therefore meaningful to investigate the redundancy in GANs and explore an effective approach for learning efficient generators of fewer parameters and calculations.

\begin{figure*}
	\centering
	\includegraphics[width=0.85\textwidth]{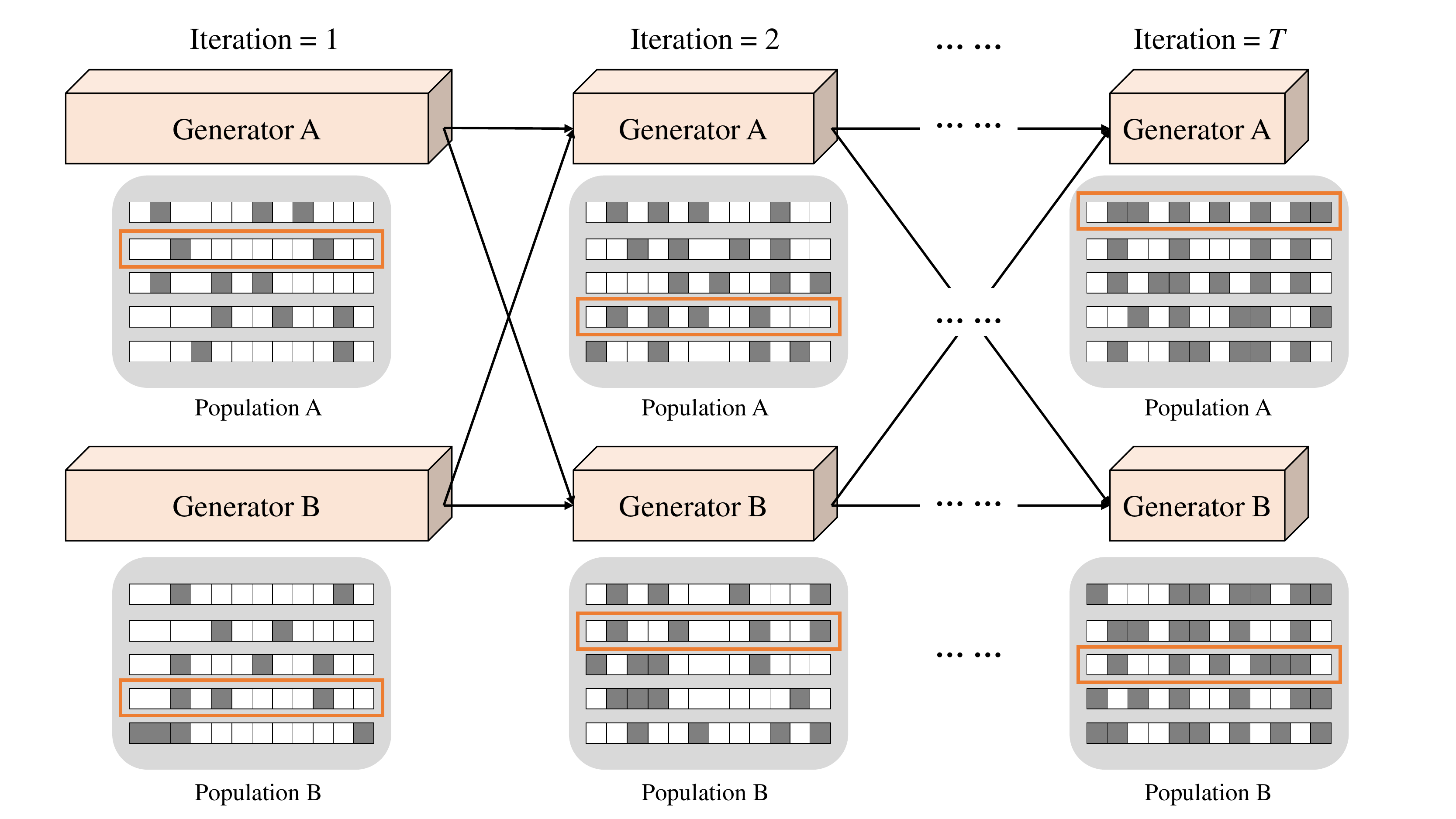}
	\vspace{0.0em}
	\caption{The diagram of the proposed co-evolutionary method for learning efficient generators. Wherein, filters in generators are represented as binary strings, and two populations are maintained for iteratively eliminating redundant convolution filters in each generator. The portable generator will be reconstructed from the best individual (red rectangle) in each population.}
	\vspace{-1.0em}
	\label{Fig:digram}
\end{figure*}

To this end, we develop a co-evolutionary algorithm to learn efficient architectures for both generators in a synergistic environment as illustrated in Figure~\ref{Fig:digram}. Wherein, convolution filters in pre-trained GANs are encoded as a binary string, so that compression and speed-up task can be converted into a binary programming problem. Generators in an image to image translation task could have different redundancies, \eg the generator for converting cityscape images to pixel images would have more redundant than that of the generator for converting pixel images to cityscape images. Two populations are therefore maintained for these two generator networks in the unpaired image translation task, respectively. The fitness of each individual is calculated in terms of the model size and a discriminator-aware loss from GANs. These two populations are updated alternatively by exploiting the best individuals in the previous iteration for obtaining portable architectures of satisfactory performance. Extensive experiments on benchmark datasets show that the proposed co-evolutionary algorithm  can effectively compress two generators simultaneously while preserving the quality of transformed images. The compressed generator has less than $1/4$ parameters compared to the original one while maintaining the image translation performance.

The rest of this paper is organized as follows. Section~\ref{sec:related} investigates related works on GANs and model compression methods. Section~\ref{sec:method} proposes the co-evolutionary approach for removing redundant filters in pre-trained GANs. Section~\ref{sec:exp} shows the experimental results conducted on benchmark datasets and models, and Section~\ref{sec:conclu} concludes the paper.

\section{Related Works}\label{sec:related}
Our goal is to reconstruct compact and efficient generators for image translation based on pretrained generator networks. There are a number of works proposed for image translation with GANs and model compression for compact deep neural networks, which will be reviewed respectively.
\subsection{GANs for Image Translation }
Generative adversarial networks have achieved impressive performance on the image translation task by adopting a discriminator network to refine the generator. Isola~\etal~\cite{pix2pix} introduced generative adversarial networks and $L_1$ loss to address the issue of paired image translation. Wang~\etal~\cite{pix2pixHD} presented a coarse-to-fine generator and a multi-scale discriminator to generate high-resolution images. Zhu~\etal~\cite{cycleGAN} implemented the unpaired image-to-image translation by utilizing two opposite domain transfer generators and a well-designed cycle loss. Similarly, Kim~\etal~\cite{discoGAN} and Yi~\etal~\cite{dualGAN} also adopted the cycle strategy to solve the unpaired image translation problem. Choi~\etal~\cite{starGAN} extended the two-domain translation task to a multi-domain image translation problem.
 
In fact, those carefully designed generator networks contain tremendous parameters and demand huge computation cost which cannot be efficiently launched on mobile devices, \eg phones and cameras. Thus, we are motivated to explore a compression method to reduce their parameters and computational complexities.

\subsection{Model Compression}
To learn compact and efficient networks from pretrained models, Denton~\etal~\cite{SVD} utilized singular value decomposition (SVD) to achieve the low-rank approximation for parameters in fully-connected layers. Chen~\etal~\cite{HashDCT} used a hash function and represented weights in the same hash bucket with a single parameter. Han~\etal~\cite{pruning15} removed unimportant parameters in pre-trained neural networks and further~\cite{pruning} utilized quantization, Huffman encoding, and sparse row format for obtaining extremely compact models. Luo~\etal~\cite{ThiNet} removed redundant filters and replaced the fully-connected layers by global average pooling (GAP) layers. Vanhouche~\etal~\cite{quan1} explored deep neural networks with 8-bit integer values to replace original models with 32-bit floating values to achieve the compression and speed-up directly. Courbariaux and Bengio~\cite{binary} explored neural networks with binary weights and activations. Restgari~\etal~\cite{XNORNet} further incorporated binary convolutions into the modern neural architecture to achieve higher performance.

Although these aforementioned methods achieved considerable speed-up and compression ratios on several benchmark models, most of them are developed for recognition tasks such as image classification and object detection, which cannot be straightforwardly applied for generator networks. Actually, GANs consist of a generator and a discriminator, whose outputs are images of high dimension and complex structures. Therefore, we shall develop effective approach for compressing GANs and preserving the visual quality of generated images.

\section{Co-Evolution for Efficient Generators}\label{sec:method}
Here we first briefly introduce the CycleGAN~\cite{cycleGAN} for unpaired image-to-image translation, which is the state-of-the-art method for learning the correspondence between two distributions using unpaired data, and then propose a novel co-evolutionary algorithm for simultaneously compressing its two generators.

\subsection{Modeling Redundancy in Generators}
Formally, given the training dataset from two different domains (\eg Zebra and Horse) $X = \{x_i\}_{i=1}^m$ and $Y = \{y_i\}_{i=1}^n$  with $m$ and $n$ images, respectively. The data distributions of these two domains are denoted as $x\sim p_{data}(x)$ and $y\sim p_{data}(y)$. The goal of CycleGAN is to learn two mappings simultaneously, \ie $G_1: X \rightarrow \mathcal{Y}$ and $G_2: Y \rightarrow X$. For the first mapping $G_1$ and its discriminator $D_1$, the corresponding objective function can be mathematically formulated as
\begin{equation}
\begin{aligned}
\mathcal{L}_{GAN}(G_1,&D_1,X,Y) = \mathbb{E}_{y\sim p_{data}(y)} \left[\log D_1(y)\right] \\
&+\mathbb{E}_{x\sim p_{data}(x)}\left[\log(1-D_1(G(x)))\right],
\end{aligned}
\label{Fcn:GAN}
\end{equation}
wherein, the functionality of the generator $G_1$ is to generate images $G_1(x)$ which looks similar to those from the other domain $Y$. The discriminator network $D_1$ is to distinguish between images generated by $G_1$ and real images in $Y$. The generator $G_1$ aims to minimize Eq.~\ref{Fcn:GAN} while the discriminator tries to maximize it, \ie
\begin{equation}
\min_{G_1}\max_{D_1}\mathcal{L}_{GAN}(G_1,D_1,X,Y),
\label{Fcn:GANtrain}
\end{equation} 
and the entire objective of the CycleGAN is
\begin{equation}
\begin{aligned}
\mathcal{L}(G_1,G_2,D_1,D_2) = &\mathcal{L}_{GAN}(G_1,D_1,X,Y)+\\
\mathcal{L}_{GAN}(G_2,&D_2,Y,X)+\lambda\mathcal{L}_{cyc}(G_1,G_2),
\end{aligned}
\label{Fcn:cycleGAN}
\end{equation}
where $\mathcal{L}_{cyc}$ is the cycle consistency loss, and $\lambda$ is the hyper-parameter for seeking the tradeoff between the generation ability and the cycle consistency. It is obvious that, the training of CycleGAN is a more complex procedure than those of recognition tasks, \eg classification~\cite{AlexNet,VGGnet,ResNet} and detection~\cite{fasterRCNN,SSD}.

In addition, although GANs perform well on image style transfer, most of generators in these models are well designed with considerable parameters and FLOPs, which are usually unaffordable on mobile devices. In addition, by analyzing Eq.~\ref{Fcn:cycleGAN}, we can find that there are two major differences between the task for compressing image classification or detection models and the task compressing generative networks for image style transfer: 1) discriminator network will be dropped after training the entire generative network, which does not need to be compact; 2) output results of GANs are of high-dimensional, and it is hard to quantitatively evaluate the generated images. We aim to explore effective methods for discovering redundant parameters and compressing original GANs to obtain efficient models.

A straightforward method for reducing complexities of GANs can be directly borrowed from the conventional pruning methods~\cite{pruning,CNNpackNIPS} for minimizing the reconstruction error on the output data, which can be formulated as a generator-aware loss function, \ie
\begin{equation}
\mathcal{L}_{GenA} = \frac{1}{m}\sum_{i=1}^m||G_1(x_i)-\hat{G}_1(x_i)||_2^2,
\label{Fcn:obj1} 
\end{equation}
where $||\cdot||_2$ is the conventional $\ell_2$-norm for calculating the difference between generated images using generators before and after compressing, and $\hat{G}_1$ is the compressed generator. 

Admittedly, minimizing Eq.~\ref{Fcn:obj1} can encourage images generated using $\hat{G}_1$ similar to those generated by $G_1$, but it is not closely related to the style transfer task. In fact, we cannot use an appearance loss to accurately measure the difference between two styles. For instance, a horse with eight or five black and white stripes can be both considered as successful transformations in the image translation task. Therefore, optimizing Eq.~\ref{Fcn:obj1} would not precisely excavate redundancy in the generator network $G_1$.

Although the discriminator network $D$ will be abandoned after the training procedure of Eq.~\ref{Fcn:GANtrain}, it contains important information for distinguishing images from different domains. Thus, we propose to minimize the following discriminator-aware objective function for learning the compressed generator network:
\begin{equation}
\mathcal{L}_{DisA} = \frac{1}{m}\sum_{i=1}^m||D_1(G_1(x_i))-D_1(\hat{G}_1(x_i))||_2^2,
\label{Fcn:obj2}
\end{equation}
where $D_1$ is the discriminator in the original network, which captures the style information in the target domain \wrt the training dataset $\mathcal{Y}$. Compared with Eq.~\ref{Fcn:obj1}, the above function does not force outputs of original generator and compressed generator to be similar but measures the style discrepancy of generated images through these two generators using the pre-trained discriminator, which is a more appropriate goal for efficient GANs. We will further investigate the difference between Eq.~\ref{Fcn:obj1} and Eq.~\ref{Fcn:obj2} on performance of GANs in the experiment part.

In addition, the cycle consistency should also be considered for maintaining the capacity of generators, \ie
\begin{equation}
\mathcal{L}_{cyc} = \frac{1}{m}\sum_{i=1}^m||G_2(\hat{G}_1(x_i))-x_i||_2^2.
\label{Fcn:obj3}
\end{equation}
Thus, the objective for compressing the first generator $G_1$ (\eg, horse to zebra) in CycleGAN can be written as
\begin{equation}
\hat{G}_1 = \arg\min_{G_1} \mathcal{N}(G_1)+\gamma\left(\mathcal{L}_{DisA}+\lambda\mathcal{L}_{cyc}\right),
\label{Fcn:comp1}
\end{equation}
where $\mathcal{N}(\cdot)$ counts the number of parameters in neural networks, and $\gamma$ is the hyper-parameter for balancing the performance of $\hat{G}_1$ and the compression ratio. 

Besides the objectives discussed above, there is another important issue should be taken into consideration during the compression procedure of GANs. In general, two generators in the CycleGAN have the same architectures and numbers of convolution filters with the similar capacity for conducting the image-to-image translation task. Only minimizing one generator will make the entire system of CycleGAN unstable, and thus we propose to simultaneously compress these two generators, \ie
\begin{equation}
\begin{aligned}
\hat{G}_1,\hat{G}_2& = \arg\min_{G_1,G_2} \mathcal{N}(G_1)+\mathcal{N}(G_2)\\
+&\gamma\left(\mathcal{L}_{DisA}(G_1,D_1)+\lambda\mathcal{L}_{cyc}(G_1,G_2,X)\right)\\
+&\gamma\left(\mathcal{L}_{DisA}(G_2,D_2)+\lambda\mathcal{L}_{cyc}(G_2,G_1,Y)\right).
\end{aligned}
\label{Fcn:comp}
\end{equation}
which can additionally provide two portable generators at the same time for saving the computing resource.

\subsection{Co-Evolutionary Compression}
Considering that we cannot accurately estimate the impact of each filter on the final loss according to its output in the given convolutional layer, and functionalities of different filters are interacted, we apply the evolutionary algorithm~\cite{GAgoogle,wang2018towards} to encode all convolution filters as binary codes. In addition, there are two variables in Eq.~\ref{Fcn:comp}, \ie $G_1$ and $G_2$, which have their own tasks for learning two different mappings, we thus develop a co-evolutionary approach utilizing Genetic Algorithm (GA~\cite{deb2002fast}) for compressing CycleGAN. Note that, other evolutionary algorithms such as simulated annealing~\cite{SA} and PSO~\cite{PSO} can also been applied similarly. 

\textbf{Updating $G_1$:} In practice, the filter pruning task will be regarded as a binary programming problem and the generator $G_1$ will be correspondingly represented as a binary string, \ie individual $\mathbf{p}$. Wherein, each bit is assigned to a convolution filter in the given network, \ie
\begin{equation}
\label{Fcn:individual}
B_l^{(n,:,:,:)}=\left\{
\begin{aligned}
0, &\;\; ~\text{if} ~\mathbf{p}_l(n)=0, \\
1, &\;\;\;\; ~\text{otherwise} ~,
\end{aligned}
\right.
\end{equation}
where $\mathbf{p}_l$ denotes the state of convolution filters in the $l$-th layer in $G_1$. $\mathbf{p}_l(n)=0$ means discarding the $n$-th filter in the $l$-th convolutional layer, otherwise retaining. The number of filters is about tens of thousands in conventional neural networks~\cite{ResNet,cycleGAN}, and the length of $\mathbf{p}$ for $l$ convolutional layers is tolerable. Since convolution filters in different layers are of various sizes, which has different memory usage and computational complexities, we utilize the following function to reformulate $\mathcal{N}(\cdot)$ in Eq.~\ref{Fcn:comp}: 
\begin{equation}
\mathcal{N}(\mathbf{p}) = \frac{\sum_{l=1}^{L}\left(\| \mathbf{p}_{l-1} \|_1\cdot \| \mathbf{p}_l\|_1 \cdot H_l\cdot W_l  \right)} {\sum_{l=1}^{L}(N_l\cdot C_l\cdot H_l\cdot W_l)},
\label{Fcn:Num}
\end{equation} 
which assigns convolution filters of more weights with higher importance. Wherein, $\| \mathbf{p}_{l-1} \|_1$ is the number of filters in the $l-1$-th layer, \ie the channel number in the $l$-th layer, $N_l$, $C_l$, $H_l$ and $W_l$ are the number of filters, the number of channels and height and width of filters in the $l$-th convolutional layer in $G_1$ respectively. Besides memory usage, Eq.~\ref{Fcn:Num} also takes FLOPs into consideration since a convolution filter with more weights, \ie $C_l\times H_l\times W_l$ usually involves more multiplications in GANs.

\begin{algorithm}[t]
	\caption{Co-Evolutionary compression for GANs.}
	\label{Alg:main}
	\begin{algorithmic}[1]
		\REQUIRE Training set $X=\{x_i\}_{i=1}^m$ and $Y=\{y_i\}_{i=1}^n$. The pre-trained GAN with two generators and discriminators, $G_1$, $G_2$, $D_1$, and $D_2$, parameters: $K$, $T$, $\lambda$, $\gamma$, and learning rates, \etc.
		\STATE Initialize two populations $P_0$ and $Q_0$ \wrt $G_1$ and $G_2$ with $K$ individuals, respectively;
		\STATE Select the best individuals $\hat{\mathbf{p}}^{(0)}$ and $\hat{\mathbf{q}}^{(0)}$;
		\FOR{$t = 1$ to $T$}
		\STATE Calculate the fitness of each individual in $P_{t}$:\\
		$\mathcal{F}(\mathbf{p}^{(t)}) \leftarrow [\mathcal{N}(\mathbf{p}^{(t)})+\gamma(\mathcal{L}_{DisA}(\mathbf{p}^{(t)},D_1,X)$\\
		$\quad\quad\quad\quad+\lambda\mathcal{L}_{cyc}(\mathbf{p}^{(t)},\hat{\mathbf{q}}^{(t-1)},X))]^{-1}$;
		\STATE Calculate the fitness of each individual in $Q_{t}$:\\
		$\mathcal{F}(\mathbf{q}^{(t)}) \leftarrow [\mathcal{N}(\mathbf{q}^{(t)})+\gamma(\mathcal{L}_{DisA}(\mathbf{q}^{(t)},D_2,Y)$\\
		$\quad\quad\quad\quad+\lambda\mathcal{L}_{cyc}(\mathbf{q}^{(t)},\hat{\mathbf{p}}^{(t-1)},Y))]^{-1}$;
		\STATE Obtain selecting probabilities (Eq.~\ref{Fcn:prob});
		\STATE Preserve the best individuals:\\
		$\quad\quad P_{t}^{(1)}\leftarrow \hat{\mathbf{p}}^{(t-1)}$, $Q_{t}^{(1)}\leftarrow \hat{\mathbf{q}}^{(t-1)}$;
		\FOR{$k = 2$ to $K$}	
		\STATE Generate a random value $s\sim[0,1]$;
		\STATE Conduct selection, crossover, and mutation for generating new individuals according to $s$;
		\ENDFOR
		\ENDFOR
		\STATE Update fitnesses of individuals in $P_{t}$ and $Q_{t}$;
		\STATE Establish two generator networks $\hat{G}_1$ and $\hat{G}_2$ by exploiting to the best individual $\hat{\mathbf{p}}^{(T)}$ and  $\hat{\mathbf{q}}^{(T)}$, respectively;
		\ENSURE Portable generator $\hat{G}_1$ and $\hat{G}_2$ after fine-tuning using the entire training set.
	\end{algorithmic}
\end{algorithm}

Then, the fitness of an individual for compressing the generator $G_1$ is defined as
\begin{equation}
\begin{aligned}
\mathcal{F}(\mathbf{p}) = &\left[\mathcal{N}(\mathbf{p})+\gamma\left(\mathcal{L}_{DisA}(\hat{G}_1,D_1) \right. \right.\\
&\quad\quad\quad\quad+\left. \left. \lambda\mathcal{L}_{cyc}(\hat{G}_1,G_2,X)\right)\right]^{-1},
\end{aligned}
\label{Fcn:fit1}
\end{equation}
where $\hat{G}_1$ is the compressed generator corresponding to the given individual $\mathbf{p}$.

After defining the calculation of fitness, GA is adopted to find the fittest individual through several evolutions. For each individual, the corresponding compressed network is fine-tuned on a subset of the training data (\eg 10\% training images randomly sampled) and take the fitness on validation set as evaluation metric. Then a probability is assigned to each individual by comparing its fitness among the individuals in the current population:
\begin{equation}
\Pr(\mathbf{p}^j) =  \left. \mathcal{F}(\mathbf{p}^j) \middle/ \sum_{k=1}^{K}\mathcal{F}(\mathbf{p}^k) \right.,
\label{Fcn:prob}
\end{equation}
where $\mathbf{p}^j$ is the $j$-th individual in the population and $K$ is the number of individuals in the population. The population in each iteration are regarded as parents, and selected according to Eq.~\ref{Fcn:prob}. The selected parents breed another population as offspring using the following operations: \textbf{selection}, \textbf{crossover}, and \textbf{mutation}~\cite{deb2002fast,wang2018towards}. 

\textbf{Updating $G_2$:} Although architectures of two generators in the CycleGAN are usually symmetrical with the same amount of convolution filters, redundancy in $G_1$ and $G_2$ can be significantly different. For instance, learning the mapping from semantic maps to streetviews is harder than that of from streetviews to semantic maps. Therefore, we utilize another population to optimize the other generator in CycleGAN, \ie $G_2$.

Similarly, we encode all convolution filters in $G_2$ to formulate a population with $K$ individuals, \ie $\mathbf{q}^1,...,\mathbf{q}^K$, and the corresponding fitness can be defined as
\begin{equation}
\begin{aligned}
\mathcal{F}(\mathbf{q}) = &\left[\mathcal{N}(\mathbf{q})+\gamma\left(\mathcal{L}_{DisA}(\hat{G}_2,D_2) \right. \right.\\
&\quad\quad\quad\quad+\left. \left. \lambda\mathcal{L}_{cyc}(\hat{G}_2,G_1,Y)\right)\right]^{-1},
\end{aligned}
\label{Fcn:fit2}
\end{equation}
which can be also optimized during the evolutionary procedure.

Moreover, it can be found in Eq.~\ref{Fcn:fit1} and Eq.~\ref{Fcn:fit2} that, the calculation of cycle consistency of each generator involves the other generator in the CycleGAN. Thus, two populations are alternatively updated to simultaneously optimize $G_1$ and $G_2$. In specific, for the $t$-th iteration, we first obtain the best individual $\mathbf{p}^{(t)}$ of $G_1$ utilizing the best individual of $G_2$ in the previous iteration $\mathbf{q}^{(t-1)}$, and then utilize it to calculate the fitness of $G_2$. In addition, the best individual preserving strategy is also adopted to increase the robustness of the evolutionary algorithm. The detailed procedure for learning portable GANs using the proposed method is summaries in Algorithm~\ref{Alg:main}.

\begin{figure*}[t]
	\centering
	\vspace{-1.5em}
	\setlength{\tabcolsep}{1pt}
	\begin{tabular}{ccccc}
		\small Input Images & \small Original Results & \small $\gamma = 0.1$ & \small $\gamma = 1$ & \small $\gamma = 10$ \\
		\includegraphics[width=0.19\linewidth]{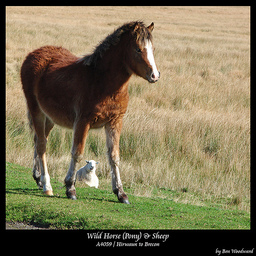} & 
		\includegraphics[width=0.19\linewidth]{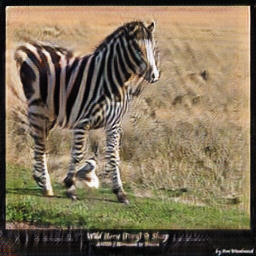} & 
		\includegraphics[width=0.19\linewidth]{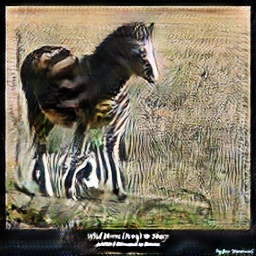} & 
		\includegraphics[width=0.19\linewidth]{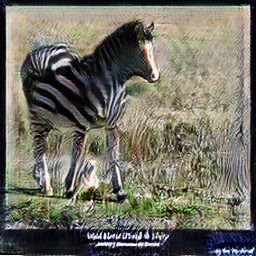} & 
		\includegraphics[width=0.19\linewidth]{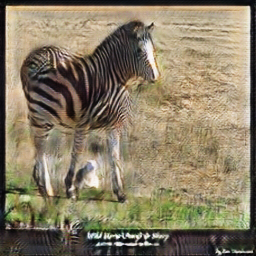} \\
		\small Model size & \small 43.42\emph{MB} & \small 7.21\emph{MB} & \small 8.07\emph{MB} & \small 10.16\emph{MB}\\
		\includegraphics[width=0.19\linewidth]{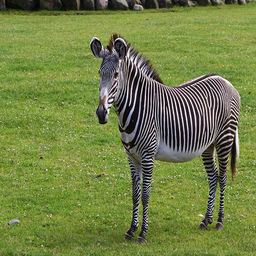} &
		\includegraphics[width=0.19\linewidth]{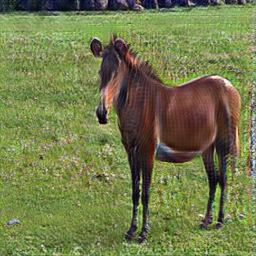} & 
		\includegraphics[width=0.19\linewidth]{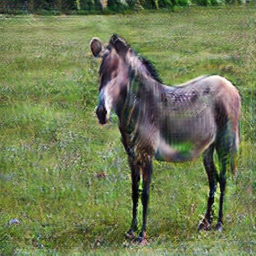} & 
		\includegraphics[width=0.19\linewidth]{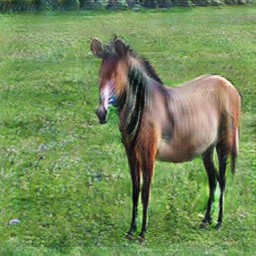} & 
		\includegraphics[width=0.19\linewidth]{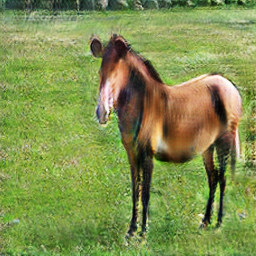} \\
		\small Model size & \small 43.42\emph{MB} & \small 7.20\emph{MB} & \small 7.85\emph{MB} & \small 10.00\emph{MB}\\
	\end{tabular}
	\label{Fig:params}
	\vspace{0.0em}
	\caption{Images generated using the generator compressed by exploiting the proposed method with different hyper-parameters. The top line and the bottom line show the results from horse to zebra, and results from zebra to horse, respectively. Two generators are compressed from an entire CycleGAN. Model sizes of different generators are provided.}
	\vspace{-1.0em}
\end{figure*}

\section{Experiments}\label{sec:exp}
In this section, we qualitatively and quantitatively evaluate the proposed discriminator aware compression method on three benchmark unpaired image translation datasets, \ie horse2zebra, summer2winter, and cityscapes. The architecture of CycleGAN is directly borrowed from their original paper~\cite{cycleGAN}. Each generator in the CycleGAN is sequentially composed of one $7\times7$ stride-1 convolutional layer, two $3\times3$ stride-2 convolutional layers, nine residual blocks~\cite{ResNet}, two $3\times3$ stride-2 transpose convolutional layers and one $7\times7$ stride-1 convolutional layer. In addition, each discriminator consists of 5 convolutional layers and one FCN~\cite{FCN} classification layer. We use the default setting in~\cite{cycleGAN} to pretrain and finetune the CycleGAN for having a fair comparison. 

\textbf{Impact of Parameters.} Our goal is to learn efficient generative network for unpaired image-to-image style transfer. As discussed in the Algorithm~\ref{Alg:main}, the objective function Eq.~\ref{Fcn:comp} for compressing GANs will be converted as the fitness calculation in the framework of the proposed co-evolutionary approach. $\lambda$ is the parameter for weighting the cycle consistency term, which is set as $10$ according to the original CycleGAN~\cite{cycleGAN}. In addition, the identity loss for maintaining the information of each domain is also applied along with the cycle consistency. The number of individuals $K$ is set as $32$, and the maximum iteration number $T$ is equal to $100$, which refer to those in~\cite{wang2018towards}.

Then, we further investigate the trade-off between the compression ratio and performance of compressed generative networks according to different hyper-parameter $\gamma$. It can be found in Figure~\ref{Fig:params} that, a larger $\gamma$ brings a lower compression ratio, \ie the model size is much smaller than that of the original model. However, the visual quality of the resulting images will be better for a larger $\gamma$.

As a result, we set $\gamma = 10$ to obtain the compressed model with an acceptable generative ability, images generated using the compressed network are similar to those using the original model. The compression ratios of two generators are $4.27\times$ and $4.34\times$, respectively. In addition, compression ratios of two generators are similar since the difficulty for transferring horses to zebras is also similar to that of transferring zebras to horses. 

\textbf{Ablation Study.} After investigating the trade-off between the generative ability and model size of GANs, we further conduct extensive ablation experiments to evaluate the functionality of different components in the proposed scheme.

\begin{figure*}[t]
	\centering
	\vspace{-1.0em}
	\setlength{\tabcolsep}{1pt}
	\begin{tabular}{cccccc}
		\small Input Images & \small Original Results & \small (a) & \small (b) & \small (c) & \small (d)\\
		\includegraphics[width=0.15\linewidth]{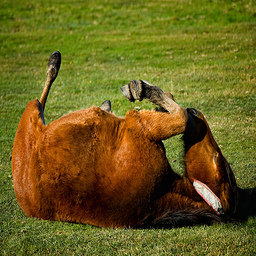} & 
		\includegraphics[width=0.15\linewidth]{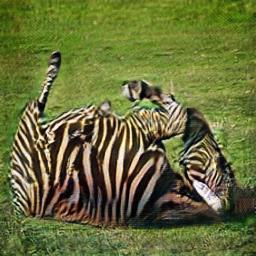} & 
		\includegraphics[width=0.15\linewidth]{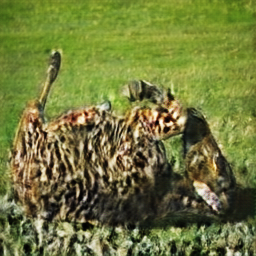} & 
		\includegraphics[width=0.15\linewidth]{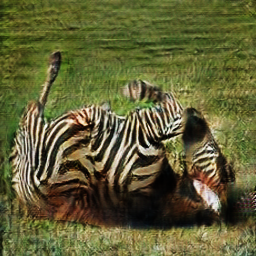} & 
		\includegraphics[width=0.15\linewidth]{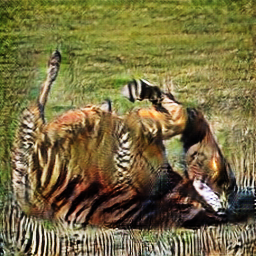} &
		\includegraphics[width=0.15\linewidth]{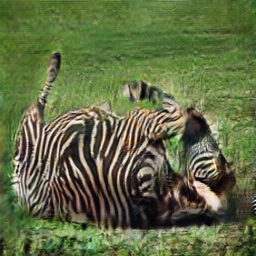}\\
		
		\includegraphics[width=0.15\linewidth]{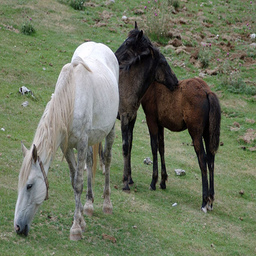} & 
		\includegraphics[width=0.15\linewidth]{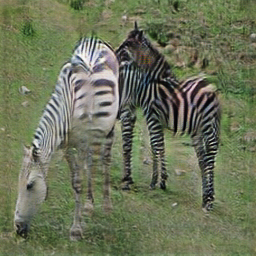} & 
		\includegraphics[width=0.15\linewidth]{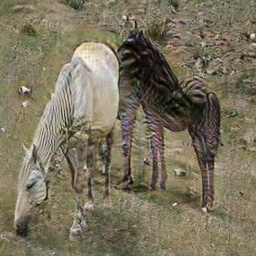} & 
		\includegraphics[width=0.15\linewidth]{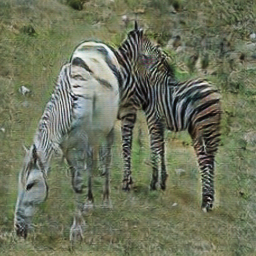} & 
		\includegraphics[width=0.15\linewidth]{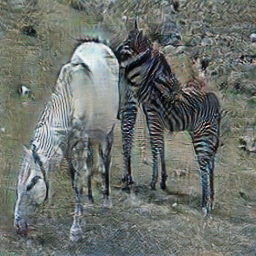} &
		\includegraphics[width=0.15\linewidth]{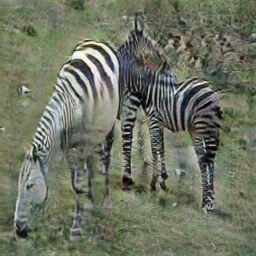}\\
		\includegraphics[width=0.15\linewidth]{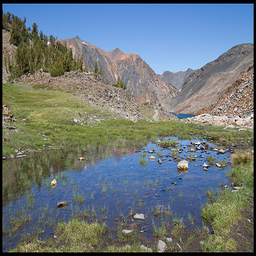} & 
		\includegraphics[width=0.15\linewidth]{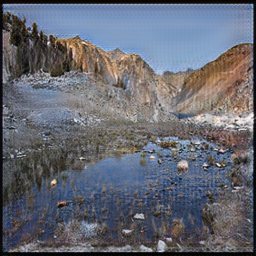} & 
		\includegraphics[width=0.15\linewidth]{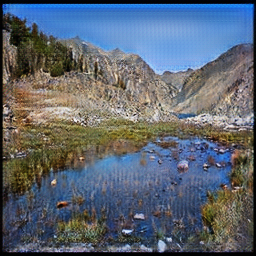} & 
		\includegraphics[width=0.15\linewidth]{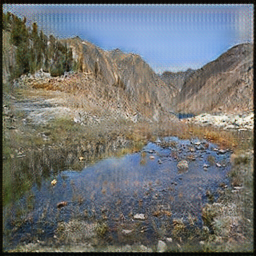} & 
		\includegraphics[width=0.15\linewidth]{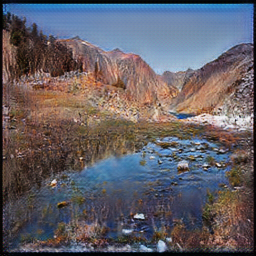} &
		\includegraphics[width=0.15\linewidth]{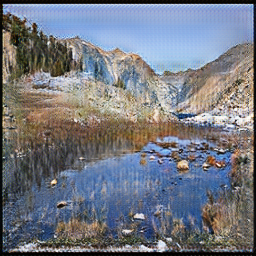}\\
		\includegraphics[width=0.15\linewidth]{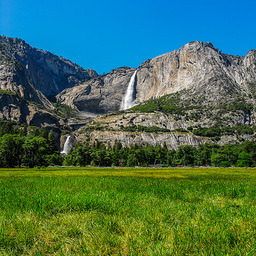} & 
		\includegraphics[width=0.15\linewidth]{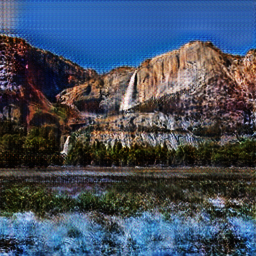} & 
		\includegraphics[width=0.15\linewidth]{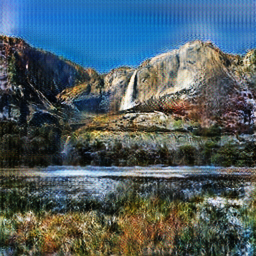} & 
		\includegraphics[width=0.15\linewidth]{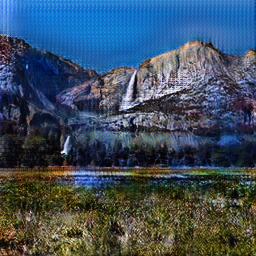} & 
		\includegraphics[width=0.15\linewidth]{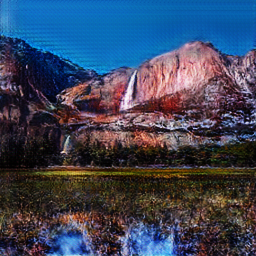} &
		\includegraphics[width=0.15\linewidth]{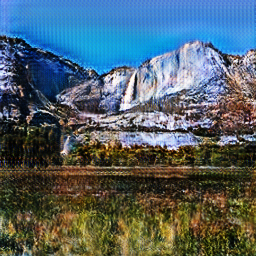}
	\end{tabular}
	\caption{The generated images on horse2zebra and summer2winter datasets using different methods and strategies. The first two columns illustrate the input images and images generated by the original CycleGAN. Images in (a) are generated by by compressed generators using the conventional ThiNet for filter pruning, while (b) by evolutionary approach for compressing two generators separately using $\mathcal{L}_{DisA}$, (c) by co-evolutionary method using $\mathcal{L}_{GenA}$, (d) by co-evolutionary method using $\mathcal{L}_{DisA}$.}
	%
	\label{Fig:ablation}
	\vspace{-1.0em}
\end{figure*}

A co-evolutionary approach for iteratively compressing two generators in the CycleGAN was developed in Section~\ref{sec:method}, which involves two populations for obtaining generative models with higher performance. Thus, we first compare the results using the evolutionary algorithm to compress two generators separately and those from the proposed co-evolutionary algorithm, as shown in Figure~\ref{Fig:ablation}(b) and Figure~\ref{Fig:ablation}(d), respectively. 

In order to have a fair comparison, we tune the hyper-parameter to obtain compressed network with the similar model size, \eg the generator using the proposed method is $10.16$\emph{MB} on the horse2zebra task. It is clear that, the proposed co-evolutionary approach obtained images with higher visual quality, \eg clear zebra pattern and more white mountains, since the proposed method can simultaneously investigate the redundancy in both two generators. In addition, we can obtain two efficient and effective generators at the same time, which is much more flexible than the scheme for compressing them separately.

We then compared the performance of the proposed two loss functions for evaluating the capacity of compressed GANs, \ie the generator-aware loss and the discriminator-aware loss. The results of compressed models under the generator-aware constraint are shown in Figure~\ref{Fig:ablation}(c). It is obvious that, the generated images using the generator-aware loss $\mathcal{L}_{GenA}$ are worse than those using the discriminator-aware loss, since the style information cannot be easily captured by the reconstruction error. For example, the difference between horses and zebras are only exist on the body of horses, the overall difference between input images and Figure~\ref{Fig:ablation}(c) are not significant.

\textbf{Comparison with Conventional Pruning Method.} In contrast to conventional method for pruning redundant convolution filters in pre-trained deep neural networks, the proposed method introduces a discriminator aware loss, \ie Eq.~\ref{Fcn:obj2} to recognize useless filters for conducting the image style transfer task. Therefore, we then compare the proposed method with the state-of-the-art filter pruning method, namely, ThiNet~\cite{ThiNet}, which minimizes the reconstruction error of output features. Similarly, we also tuned the hyper-parameters in ThiNet to ensure that the model size ($10.88$\emph{MB}) of resulting generator of ThiNet is similar to that of using the proposed method.

It can be found in Figure~\ref{Fig:ablation}(a), images generated through a generator compressed by ThiNet for a similar amount of parameters cannot capture the style information in the target domain, \eg the generated zebra images are fundamentally different to those of original model and compressed model using the proposed method, as shown in Figure~\ref{Fig:ablation}(d). In fact, the conventional filter pruning method has the similar assumption to that of the generator-aware loss, which obtained similar but worse results as those in Figure~\ref{Fig:ablation}(c), which is not suitable to conduct the compression task for the unpaired image translation. In addition, we also compare the proposed method with other filter pruning methods, namely, network trimming~\cite{trimming} and slimming~\cite{slimming}, which obtained similar results to those of ThiNet and can be found in the supplementary materials.

\begin{figure*}[t]
	\centering
	\vspace{-1.5em}
	\includegraphics[width=0.70\textwidth]{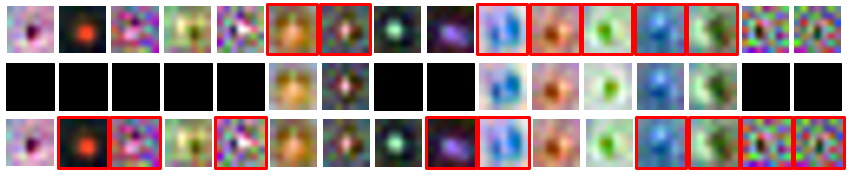}\\	
	\caption{Filter visualization results. From top to bottom: the original filters with red rectangles selecting the remained filters by the proposed method,  filters after fine-tuning, and the original filters with red rectangles selecting the filters remained by ThiNet.}
	\label{Fig:filter}
	\vspace{0.0em}
\end{figure*}

\textbf{Filter Visualization.} Since the evolutionary algorithm can globally discover the most beneficial filters for given task, it is necessary to see what filters are recognized as redundant and what filters are essential for generator. Thus, we visualize the first several filters in the first convolutional layer of CycleGAN on the horse2zebra dataset as shown in Figure~\ref{Fig:filter}. Interestingly, the discarded filters by our method are not only with small norms but may also have big values, which is significantly different from the results of the conventional filter pruning method, \ie ThiNet~\cite{ThiNet}. Actually, the weights in filters for extracting color and texture information can be very small.

It can be found in Figure~\ref{Fig:filter}, the proposed method retains filters with more distinct structures, which are beneficial for maintaining an acceptable performance of the generator network. Furthermore, filters after fine-tuning do not have significant changes, which demonstrates importance and functionality of these convolution filters for conducting the subsequent image-to-image translation task.

\begin{table}[h]
	\vspace{-0.5em}
	\caption{Statistics of compressed generators.}
	\label{Tab:comp}
	\small
	\begin{tabular}{c|c|c|c|c}
		\hline
		Task & Memory & $r_c$ & FLOPs & $r_s$\\
		\hline\hline
		horse2zebra & $10.16$\emph{MB} & $4.27\times$ &  $13,448\emph{M}$ &$4.23\times$ \\
		zebra2horse & $10.00$\emph{MB}  & $4.34\times$ &  $13,060\emph{M}$ &$4.35\times$ \\
		\hline
		summer2winter & $7.98$\emph{MB}& $5.44\times$  &$11,064\emph{M}$ &$5.14\times$ \\
		winter2summer & $7.61$\emph{MB} & $5.70\times$  &$10,994\emph{M}$ &$5.17\times$ \\
		\hline
		cityscapes-A2B & $8.98$\emph{MB}& $4.84\times$  &$12,977\emph{M}$ &$4.38\times$ \\
		cityscapes-B2A & $12.26$\emph{MB}& $3.54\times$ &$16,445\emph{M}$ &$3.46\times$ \\
		\hline
	\end{tabular}
	\vspace{-0.5em}
\end{table}

\textbf{Detailed Compression results.} Moreover, detailed results of the six generators trained on three datasets, \ie horse2zebra, summer2winter, and cityscapes, are illustrated in Table~\ref{Tab:comp}, $r_c$ and $r_s$ are compression rates for model size and FLOPs respectively. It is obvious that, the proposed co-evolutionary method can effectively remove redundant filters in pre-trained GANs and obtain efficient generators. Additionally, we can obtain two efficient generators from CycleGAN~\cite{cycleGAN} for conducting the unpaired image-to-image translation task by simultaneously explore their redundancy. Furthermore, there are some interesting phenomenons in Table~\ref{Tab:comp}, \ie the generator for more difficult transformation task tend to have less redundancy. For instance, since the task for transferring semantic map to streetview is more difficult than that of transferring streetview to semantic map. The model size (12.26\emph{MB}) of the second compressed generator (\ie cityscapes-B2A) is much larger than that (8.98\emph{MB}) of the first generator (\ie cityscapes-A2B) for transferring streetview to semantic map, which demonstrates the superiority of the proposed co-evolutionary approach for compressing GANs and provides some guidance for designing GANs for various tasks. In addition, detailed statics of two generators in the compressed CycleGAN and more visualization results on three benchmark datasets generated of compressed models using the proposed method can be found in the supplementary materials.

\textbf{Runtime. } The proposed method directly removes redundant filters and produces efficient GANs. Thus, the compressed model does not require other additional support (\eg sparse matrices and Huffman encoding~\cite{pruning}) for realizing the network speed-up. We then compared runtimes for processing images using original and compressed models.
In practice, the averaged runtime of the original model for processing one image is about $2,260$ms using an Intel Xeon E5-2690 CPU. In contrast, the runtime of the compressed model with a 10.16\emph{MB} model size (\ie the first line in Table~\ref{Tab:comp}) is about $730$ms, which achieved an about $3.1\times$ speed-up, which is lower than that of the theoretical speed-up ratio ($4.23\times$) due to the costs of incurred by data transmission, ReLU, \etc. The demo code for verifying the proposed method can be found in our supplementary materials.

\begin{table*}[t]
	\centering
	\vspace{-0.5em}
	\caption{FCN scores of different generators calculated on the cityscapes dataset. }
	\label{Tab:FCN}
	\small
	\begin{tabular}{c||c|c|c|c}
		\hline
		Method & Memory & Mean Pixel Acc & Mean Class Acc. & Mean class IoU \\
		\hline\hline
		Original~\cite{cycleGAN}  &$43.42$\emph{MB}  &0.538 &0.172 &0.121 \\
		ThiNet~\cite{ThiNet} &$10.88$\emph{MB}  &0.218 &0.089 &0.054 \\
		Ours  &$12.26$\emph{MB}  &0.542 &0.212 &0.131 \\
		\hline
	\end{tabular}
	\vspace{-1.5 em}
\end{table*}

\textbf{Quantitative Evaluation.}
Besides the above experiments, we also conduct the quantitative evaluation of the proposed method. In order to evaluate the quality of compressed generators the ``FCN-score''~\cite{pix2pix} is utilized on images generated from semantic maps to cityscape images. In practice, a pre-trained FCN-8s network~\cite{FCN} on the cityscapes dataset is exploited for conducting the semantic segmentation experiments and detailed results are shown in Table~\ref{Tab:FCN}. Measurements for the segmentation experiment are per-pixel accuracy, per-class accuracy and mean class IOU. 
It is obvious that, the proposed method obtained better results compared with the conventional ThiNet~\cite{ThiNet} for pruning convolution filters, which are slightly higher results than those of using the original generator, since we can effectively remove useless filters to establish generative models perform well on discriminator networks. In addition, the segmentation results are shown in our supplementary materials.

On datasets of horse2zebra and summer2winter, Fr\'{e}chet Inception Distance 
(FID) is adopted to evaluate the results of the proposed method as shown in Table \ref{Tab:FID}. The results of proposed method are close to the original CycleGAN\cite{cycleGAN}, obviously better than the weight based pruning method.
\begin{table}[h]
	\vspace{-0.5em}
	\centering
	\caption{Comparision of FID scores.}
	\label{Tab:FID}
	\small
	\begin{tabular}{c|c|c|c}
		\hline
		FID & Original\cite{cycleGAN} & ThiNet\cite{ThiNet} & Ours \\
		\hline\hline
		horse2zebra & $74.04$ & $189.28$ &  $96.15$  \\
		zebra2horse & $148.81$  & $184.88$ &  $157.90$  \\
		\hline
		summer2winter & $79.12$ & $81.06$ &  $78.58$ \\
		winter2summer & $73.31$ & $80.17$ &  $79.16$ \\
		\hline
	\end{tabular}
	\vspace{-1.5em}
\end{table}
\section{Conclusion}\label{sec:conclu}
This paper studies the model compression and speed-up problem of generative networks for unpaired image-to-image style translation. A novel co-evolutionary scheme is developed for simultaneously pruning redundant filters in both two generators. Two portable generator networks will be effectively obtained during the procedure of the genetic algorithm. Experiments conducted on benchmark datasets and generative models demonstrate that the proposed co-evolutionary compression algorithm can fully excavate redundancy in GANs and achieve considerable compression and speed-up ratios. In addition, images generated using the compressed generator also maintain the style information with high visual quality, which can be directly applied on any off-the-shelf platforms.

\textbf{Acknowledgments:} We thank Dr.~Gang Niu for the insightful discussion. Chang Xu was supported by the Australian Research Council under Project DE180101438.

{
\clearpage	
\small
\bibliographystyle{ieee}
\bibliography{ref}
}

\end{document}